\documentclass[sn-basic,Numbered]{sn-jnl}

\usepackage{graphicx}\usepackage{multirow}\usepackage{amsmath,amssymb,amsfonts}\usepackage{amsthm}\usepackage{mathrsfs}\usepackage[title]{appendix}\usepackage{xcolor}\usepackage{textcomp}\usepackage{manyfoot}\usepackage{booktabs}

\usepackage{amssymb}
\usepackage{xcolor}
\usepackage{hyperref}
\usepackage{url}

\usepackage[switch]{lineno}
\usepackage{multirow,multicol}
\usepackage{makecell}
\usepackage{colortbl}

\usepackage[final]{changes}
\definechangesauthor[name={Ahmad Pouramini}, color=blue]{R2}
\definechangesauthor[name={Ahmad Pouramini}, color=teal]{R3}

\theoremstyle{thmstyletwo}
\theoremstyle{thmstylethree}
\begin{document}

\title[Enhancing Few-Shot Transfer Learning with Optimized Multi-Task Prompt Tuning through Modular Prompt Composition]{Enhancing Few-Shot Transfer Learning with Optimized Multi-Task Prompt Tuning through Modular Prompt Composition}
\author*[1,2]{\fnm{Ahmad} \sur{Pouramini}}\email{ahmad.pouramini@ut.ac.ir}

\author[1]{\fnm{Hesham} \sur{Faili}}\email{hfaili@ut.ac.ir}
\equalcont{These authors contributed equally to this work.}

\affil*[1]{\orgdiv{School of Electrical and Computer Engineering, College of Engineering}, \orgname{University of Tehran}, \orgaddress{\street{North Kargar Street}, \city{Tehran}, \postcode{515-14395}, \state{Tehran}, \country{Iran}}}

\affil[2]{\orgdiv{Department of Computer Engineering}, \orgname{Sirjan University of Technology}, \orgaddress{\city{Sirjan}, \postcode{78137-33385}, \country{Iran}}}

\abstract{

Multi-task prompt tuning has recently gained significant attention due to its modular design and potential for parameter-efficient transfer learning across diverse domains, including natural language and vision tasks. This paper investigates strategies to enhance multi-task performance by enabling effective knowledge transfer between task-specific prompts.

We propose a novel framework that constructs each target task's prompt as a combination of shared source prompts and a task-specific private prompt. Several methods for integrating these components are presented and compared, with a detailed analysis of the respective roles and contributions of source and private prompts. Based on this analysis, we introduce flexible configurations that adaptively balance shared and task-specific knowledge to optimize performance.

Empirical results demonstrate consistent improvements in both accuracy and robustness over standard prompt tuning and related baselines. Our approach particularly excels in few-shot scenarios, achieving superior performance on the GLUE benchmark and other tasks, while requiring substantially less training data. These findings highlight the effectiveness of our method for data-efficient, multi-task learning.
}


\maketitle

\section{Introduction}

Transfer learning has become the dominant paradigm in natural language processing (NLP) and computer vision \cite{concept_transfer}. The prevailing approach involves pretraining a model on a large corpus of unlabeled data and then fine-tuning it on task-specific datasets. While this strategy has led to strong performance across many tasks, it suffers from well-known limitations such as \textit{catastrophic forgetting} \cite{concept-catas} and \textit{negative interference} \cite{concept-negtrans}, especially in multi-task and continual learning settings.

To address these issues, the concept of \textit{modular deep learning} has gained momentum \cite{modular}. In modular architectures, the model is decomposed into smaller, reusable components—such as adapters \cite{mahabadi}, task-specific layers \cite{concept-tasklayer2}, or embeddings—that can be selectively activated. This approach improves parameter efficiency and allows for better separation of task-specific and general knowledge, thereby encouraging positive transfer.

Prompt tuning \cite{prefix,lester2021power,gptund} offers a particularly lightweight form of modular adaptation, where a small set of learned prompt tokens is used to steer a frozen pretrained model toward a specific task. These \textit{soft prompts} effectively serve as modular parameters that encode task-relevant behavior \cite{modular,spot}. Moreover, recent studies have explored transferring or composing prompts across tasks to enhance few-shot generalization \cite{rel-attempt, rel-multi-pt, mvpt}. Yet, the effective utilization of this cross-task knowledge within a multi-task learning framework—along with a principled mechanism to balance task transfer and task specialization—has remained elusive.

\begin{figure*}[t]
	\centering
	\includegraphics[width=0.8\linewidth]{ 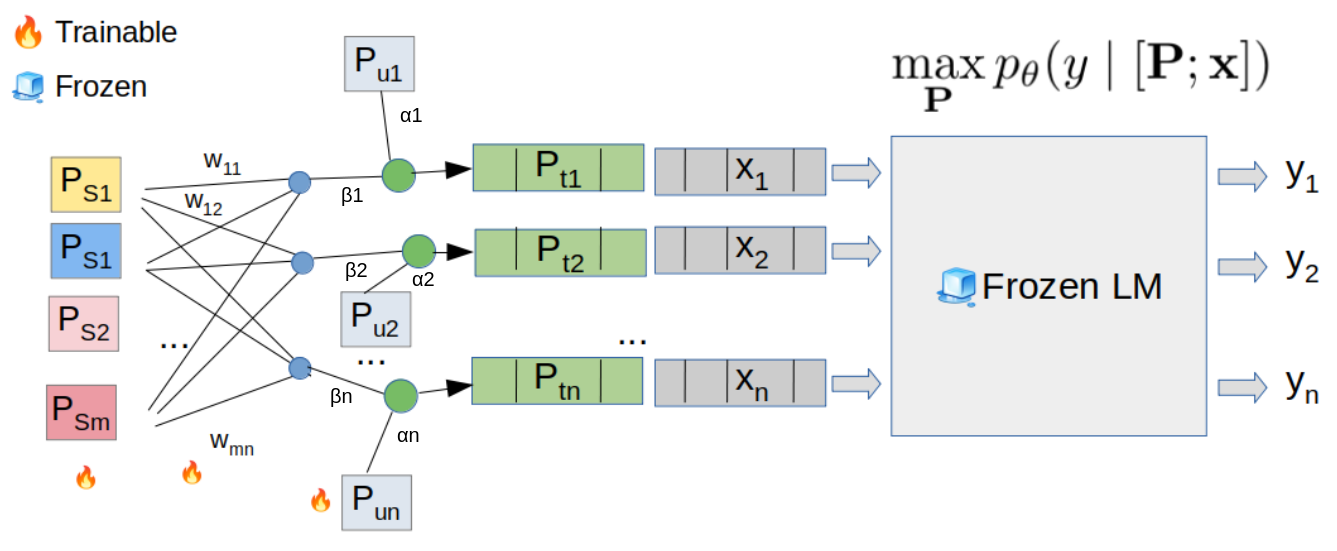}
	\caption[Overview of ComPT]{An overview of ComPT: for each target task, a target prompt is formed by combining a learned private prompt with a weighted combination of shared source prompts. The attention module assigns task-specific weights to each source prompt. The combined prompt is appended to the input and used to condition the language model.}
	\label{fig:overal}
\end{figure*}

In this paper, we introduce ComPT—a framework for Compositional Prompt Tuning tailored to multi-task few-shot learning. As illustrated in Figure~\ref{fig:overal}, ComPT constructs each task’s prompt as a weighted composition of shared source prompts and a task-specific private prompt. The combination is controlled by a learnable attention mechanism alongside adjustable gating coefficients ($\alpha$ and $\beta$), which modulate the relative contributions of shared and private components. This compositional design promotes flexible and adaptive prompt reuse, striking a balance between generalization across tasks and specialization for individual tasks. Training is conducted jointly over all tasks in a multi-task learning setup, which presents challenges such as modeling shared structures while mitigating cross-task interference. To address these challenges, ComPT jointly learns the source prompts, private prompts, and attention weights from scratch, while balancing their contributions through carefully tuned learning rates for each module.

\added[id=R2,comment={C \#2 adding vision-related works.}]
{
While our empirical evaluation focuses on NLP tasks from the GLUE benchmark, the ComPT framework is general and readily applicable to other modalities. In fact, recent work in vision and multimodal domains—such as Visual Prompt Tuning \cite{visualprompt}, MaPLe \cite{maple}, and CoOp \cite{visualprompt_cpp}—illustrates a parallel trend toward prompt-based adaptation. Our modular formulation is compatible with such architectures: shared visual prompts (e.g., image patches) and private task prompts (e.g., conditional embeddings) can be composed using the same principles described here.
}
In the following sections, we detail the architecture, composition strategies, and experimental setup. We then present results on standard benchmarks, followed by an in-depth analysis and comparison with existing approaches.

\section{Method}
This section provides a detailed description of each component of our approach. We begin by reviewing prompt tuning and explain how prompts are encoded using a prompt encoder in our framework. We then describe how task-specific and shared prompts are composed and mixed in a multi-task training setup.

\subsection{Prompt Tuning}
Prompt tuning is a lightweight and parameter-efficient method for adapting large pretrained language models (LMs) to downstream tasks. Rather than updating the model’s internal weights, prompt tuning optimizes a small set of learnable prompt embeddings that, when prepended or appended to the input, steer the model’s behavior toward task-specific outputs.

Given a template $\text{T} = \{p_1, \ldots, p_m, \mathbf{x} \}$, where $\mathbf{x} = [x_1, \ldots, x_l]$ is the input sequence and $p_i$ are prompt tokens, the model input becomes a combination of the input and the soft prompt. 

We define the soft prompt as $\mathbf{P} = [h_1, \ldots, h_m]$, where each $h_i \in \mathbb{R}^{1 \times d}$ denotes the embedding of the $i$-th prompt token, and $d$ is the embedding dimensionality of the language model. This prompt serves as a continuous conditioning signal that steers the language model toward a specific domain, task, or behavior.

The prompt is optimized by maximizing the conditional likelihood of the target output $\mathbf{y}$ given the input $\mathbf{x}$ and the prompt $\mathbf{P}$:

\begin{align}
	\max_{\mathbf{P}} \; p_\theta(\mathbf{y} \mid [\mathbf{P}; \mathbf{x}])
\end{align}

During training, only the prompt embeddings are updated, while the language model’s parameters remain frozen. To enhance the representational capacity and adaptability of the soft prompts, we apply a small neural network—referred to as the prompt encoder—to each prompt token embedding:

\begin{align}
	h_i &= \textbf{Encoder}(h_i)
\end{align}

This encoder consists of a multilayer perceptron with one hidden layer and a non-linear activation function. It serves to transform each prompt embedding into a more expressive form before passing it to the language model.

\added[id=R2,comment={C \#2 and C \#4. Extending to vision}] {
Notably, a similar mechanism exists in the vision domain, where \textbf{visual prompt tuning} prepends learnable patch tokens to the input of Vision Transformers (ViTs). In both cases—text and vision—the prompt serves the role of modulating the model’s attention and internal representations, effectively steering it toward the desired task without updating the core model parameters.
}

\subsection{Target Prompt Decomposition} \label{sec:decompose}

We consider a multi-task setup with \( \mathbf{N} \) target tasks, where each task \( t \) constructs its target prompt \( \mathbf{P}_t \) by combining a set of shared source prompts 
\[
\mathcal{S} = \{ \mathbf{P}_{s_1}, \dots, \mathbf{P}_{s_M} \}
\]
with a task-specific private prompt \( \mathbf{P}_u \). Unlike vanilla prompt tuning—which assigns an independent prompt to each task—our approach enables parameter sharing by decomposing each target prompt into a weighted combination of shared and private components.

The parameters of the shared prompts, private prompts, and their combination mechanism (e.g., attention weights) are learned jointly via backpropagation. While prompts can be optionally initialized from pretrained representations, our experiments show that learning all components from scratch—without prompt pretraining—yields strong performance.

Formally, the target prompt \( \mathbf{P}_t \) for task \( t \) is computed as follows:

\begin{align}
	\mathbf{P}_{t} = \alpha \left(  \sum_{s \in \mathcal{S}} w_{ts} \cdot \mathbf{P}_s \right) \circ \beta \, \mathbf{P}_{u}  
	\label{eq:general}
\end{align}

The weights \( w_{ts} \) are determined by an attention mechanism implemented through an attention module \( \mathcal{R} \), which dynamically assigns relevance scores to the shared source prompts \( \mathbf{P}_s \). This mechanism controls the contribution of each source prompt to a specific target task \( t \), allowing the model to selectively focus on transferable knowledge. 
\added[id=R3,comment={C \#1 Balancing knowledge transfer }]{
The scalar gates \( \alpha \) and \( \beta \) modulate the relative influence of the private and shared components in forming the final target prompt. Specifically, \( \alpha \in [0, 1] \) controls the overall weight of the private prompt \( \mathbf{P}_u \), while \( \beta \in [0, 1] \) adjusts the internal weight of the shared prompt mixture. These gates may be fixed, learned globally, or computed dynamically for each task or input instance. }

If the shared source prompts are deemed uninformative for a given task, the model can downweight their influence via the attention weights \( w_{ts} \) or the scalar \( \beta \), thereby relying more heavily on the private prompt \( \mathbf{P}_u \). This formulation enables a flexible balance between generalization and task-specific adaptation.

For composing the source prompts into the private prompt, we considered both summation and concatenation. This framework is generalizable, allowing for the exploration of various methods. Here, we cover two specific cases as follows, each with their corresponding names listed after the formula:

\begin{align}
	\mathbf{P}_{t} =  
	\begin{cases} 
		\alpha  \left( \sum_{s \in \mathcal{S}} w_{ts} \cdot \mathbf{P}_s \right) + \beta \, \mathbf{P}_{u}  & \text{if } \circ = \text{sum (SPP)} \\[0.6em]
		   \left( \sum_{s \in \mathcal{S}} w_{ts} \cdot \mathbf{P}_s \right)\mathbin\| \mathbf{P}_{u}  & \text{if } \circ = \text{concat (SCP)} \\
	\end{cases}
	\label{eq:specific}
\end{align}

In the case of SCP (Source-Contact-Private), the target prompt length is equal to the sum of the source prompt and the private prompt. Specifically, the private prompt is concatenated to the weighted combination of source prompts. The private prompt can provide task-specific instructions after the shared source prompts have encoded general, transferable semantics. This sequential arrangement allows the model to first attend to broadly useful information and then adjust its behavior based on task-specific cues. In contrast, for the summation-based variant SPP (Source-Plus-Private), both the source prompts and the private prompt must have the same length, as their combination is performed via weighted summation. In this formulation, the private prompt can serve to complement or refine the semantics encoded in the shared source prompts. We further analyze the functional impact of these combination strategies in the Discussion section (Section~\ref{sec:discuss}).

In multi-task learning, the loss for each private prompt $\mathbf{P}_{u}$ is computed only when an input sample from the corresponding task is provided to the model. In contrast, the shared source prompt $\mathbf{P}_s$ is updated at every iteration using all input samples, regardless of their task origin. To distinguish tasks, we prepend task-specific prefixes (e.g., MNLI, QNLI) to each input. This enables the shared source prompt to identify the task associated with each sample, which we found to be especially beneficial in few-shot settings compared to removing task identifiers.

\subsection{Source-Target Attention Module}
To enable dynamic composition of target prompts, we introduce an attention module $\mathcal{R}$ that assigns task-specific weights $w_{1t}, \ldots, w_{mt}$ to the $m$ shared source prompts for each target task $t$. These weights are learned end-to-end, allowing the model to softly select relevant source knowledge during prompt construction.

We explored two approaches for learning these weights: (i) direct optimization as free parameters, and (ii) sampling from a Relaxed Bernoulli distribution \cite{rb} to encourage sparsity via the reparameterization trick. Empirically, the stochastic variant yielded no clear advantage, so we adopted direct learning for simplicity and stability.
\added[id=R3,comment={C \#4 Optimization challenges}]{
\textbf{Normalization.} To reflect the relevance of source prompts to the target task, we evaluated several normalization strategies. \textit{Softmax} offers smooth gradients and introduces competition among source prompts, which helps prioritize the most relevant ones. However, as the number of prompts increases, \textit{sparsemax} \cite{sparsemax} often performs better by assigning exact zeros to irrelevant sources, thereby reducing noise and enhancing selectivity. In contrast, \textit{sigmoid} treats each source independently, allowing flexible inclusion without enforcing global competition, but often leads to overly diffuse mixtures that reduce task specificity.}

\added[id=R3,comment={C \#4 Optimization challenges}]{
Therefore, for a single source prompt, we apply sigmoid to model the degree of dependency on that source. When multiple source prompts are used, we apply softmax, and for scenarios involving more than half the number of total tasks as sources, we switch to sparsemax to promote sparsity and robustness. 
Additionally, for the gating coefficients $\alpha$ and $\beta$ in Equation~\ref{eq:specific}, which modulate the contributions of private and source prompts, we use a learnable sigmoid gating function to enable smooth control over their relative influence. }

\textbf{Fast and Slow Learning.}
Learning source-target attention requires careful regulation of adaptation speed across components. To promote effective reuse and rapid adjustment to task-specific needs, we apply a higher learning rate to the attention module. Furthermore, we prioritize learning of source prompts over private prompts to enhance knowledge sharing and cross-task generalization.

These design choices are further analyzed in the Discussion section~\ref{sec:discuss}.

\subsubsection{Inference} 

During inference, we load all shared source prompts, private prompts, and the shared attention module once. For a given task, the corresponding target prompt is constructed using Equation~\ref{eq:specific}. This target prompt is then concatenated with the input instance \( \text{x} \) and passed to the model. The model generates the output \( y \) by leveraging its internal attention mechanism to jointly process both the learned prompt and the input sequence.

In the following sections, we present the results of applying our method to a variety of tasks and provide a comprehensive comparison across different prompt composition strategies.

\section{Experiments}
\subsection{Tasks}
We assess the performance of proposed methods across various tasks, including eight tasks from the GLUE benchmark \cite{glue} and six tasks from other datasets.  These datasets are publicly available on  Hugging Face's model hub.\footnote{Retrieved from \url{https://huggingface.co/datasets/glue}}

Our evaluation involves two distinct experiments. In Experiment 1, the GLUE tasks encompass MNLI, QNLI, RTE (NLI), COLA (grammatical acceptability), SST2 (sentiment analysis), STSB (semantic similarity), MRPC, and QQP (paraphrase detection). 

For Experiment 2, we include MultiNLI, SciTail (NLI), IMDB,  Yelp-Polarity (sentiment analysis), PAWS, and MRPC (paraphrase detection). We collectively refer to this group of tasks as SET2 tasks.

\subsection{Implementation Details}
We use the T5-base language model \cite{t5} as our backbone, consistent with prior work in prompt tuning. We specifically choose a generative model like T5 because, in our setup, the label space plays a crucial role in prompt tuning and task similarity alignment. Generative models inherently incorporate label information through their output tokens, which facilitates better transfer and adaptation across related tasks. 

Prompt length plays a significant role in performance; we experimented with shorter prompts (5 tokens) and found them to be more effective than longer ones. In particular, for the SCP variant, shorter prompts led to greater stability and efficiency. Using equal lengths for the source and private prompt encoders yielded better results, ensuring a balanced contribution from both components in the concatenated prompt.

To identify an effective prompt encoder architecture, we explored various designs and activation functions. While we initially tested an LSTM-based encoder inspired by \cite{gptund}, we ultimately achieved better performance with an MLP-based encoder. Our final prompt encoder is a two-layer MLP that produces prompt embeddings of dimensionality \( d \), matching the embedding size of the language model. It consists of a fully connected layer with hidden size \( d / 2 \), followed by a GELU activation, and a second fully connected layer that outputs the final prompt embedding.

The attention module (used to compute source prompt weights) was trained with a learning rate of 0.1. For SPP, we used separate learning rates for source and private prompt encoders: 0.07 for the source prompts and 0.02 for the private prompts, enabling a balance between shared transfer and task-specific specialization. For SCP, where source and private prompts are concatenated, using a unified learning rate of 0.1 for both prompt encoders yielded the best results.

\begin{table*}[thb!]
	\centering
	\caption{Comparative performance on chosen tasks from GLUE  datasets using proposed methods in few-shot settings. The performance is computed across 3 distinct seeds.}
\label{table:res}	
	
\begin{tabular}{lcc|cccccccc}
	\hline
	method                &   k-shot & Avg.            & cola          & mnli          & mrpc          & qnli          & qqp           & rte           & sst2          & stsb          \\
	\hline
	\hline
	\multirow{3}{*}{\makecell[l]{PT \\ (baseline)}}   &                   8 & 63.35          & 35.6          & 35.3          & 75.2          & 72.4          & 76.5          & 47.7          & 81.3          & 82.9          \\
	&                  16 & 63.29          & 34.1          & 28.7          & 77.9          & 60.3          & 83.1          & 48.5          & 89.4          & 84.4          \\
	&                  32 & 70.15          & 71.5          & 45.3          & 84.3          & 52.2          & 79.5          & 48.7          & \textbf{93.1} & 86.5          \\
	\hline
\multirow{3}{*}{\makecell[l]{MST \\ (baseline)}} &          8 & 44.09          & 11.67         & 34.67         & 9.67          & 70.33         & 60.67         & 50.00         & 49.33         & 66.37         \\
&                  16 & 50.75          & 40.00         & 36.50         & 7.50          & 66.00         & 61.00         & 52.00         & 76.00         & 67.00         \\
&                  32 & 61.80          & 44.50         & 44.00         & 42.50         & 69.50         & 72.00         & 61.50         & 85.50         & 74.90         \\

	\hline
	\multirow{3}{*}{SCP} &                   8 & 74.32          & 67.0          & 58.8          & 80.3          & 87.7          & 85.2          & 57.3          & 72.9          & 85.3          \\
	&                  16 & 78.52          & 65.7          & 70.3          & 84.2          & 87.1          & 86.7          & 59.7          & 87.7          & 86.8          \\
	&                  32 & 80.40          & 68.9          & 71.6          & 83.8          & \textbf{90.9} & 87.5          & 64.0          & 90.2          & 86.2          \\
	\hline
	\multirow{3}{*}{SPP} &                   8 & 74.38          & 68.0          & 61.7          & 85.1          & 74.9          & 86.9          & 60.9          & 73.3          & 84.2          \\
	&                  16 & 79.47          & 68.3          & 72.5          & \textbf{85.8} & 82.1          & 87.5          & \textbf{65.5} & 88.3          & 85.8          \\
	&                  32 & \textbf{81.04} & \textbf{71.6} & \textbf{78.1} & 83.3          & 90.4          & 83.3          & 65.4          & 88.4          & 87.7          \\
	\hline
\end{tabular}

\end{table*}

\begin{table*}[thb!]
	\centering
	\caption{Comparative performance on tasks in SET2 using proposed methods in few-shot settings. The performance is computed across 3 distinct seeds.}
\label{table:res_lt}	

\begin{tabular}{lcc|cccccc}
	\hline
	method                &   k-shot & Avg.  & imdb  & mrpc  & multinli & paws  & scitail & yelp-polarity \\
	\hline
	\multirow{3}{*}{ \makecell[l]{PT \\ (baseline)}} 
	& 8  & 50.89 & 53.00 & 31.67 & 51.33 & 36.33 & 57.33 & 75.67 \\
	& 16 & 59.17 & 60.00 & 55.00 & 56.67 & 42.33 & 61.00 & 80.00 \\
	& 32 & 65.56 & 72.67 & 66.33 & 58.33 & 46.33 & 62.33 & 87.33 \\
	\hline
	\multirow{3}{*}{\makecell[l]{MST \\ (baseline)}} 
	& 8  & 59.50 & 67.67 & 49.00 & 57.33 & 48.33 & 59.00 & 75.67 \\
	& 16 & 67.22 & 79.67 & 58.00 & 59.67 & 49.67 & 66.33 & 90.00 \\
	& 32 & 68.89 & 80.33 & 67.00 & 61.00 & 50.00 & 67.67 & 87.33 \\
	\hline
	\multirow{3}{*}{SCP} 
	& 8  & 70.6  & 76.8  & 81.4  & 59.1  & 53.9  & 68.1  & 84.6 \\
	& 16 & 72.6  & 78.3  & 83.5  & 66.5  & 52.9  & 65.8  & 88.5 \\
	& 32 & 77.1  & 79.3  & 83.5  & 71.9  & 62.9  & \textbf{76.7} & 88.3 \\
	\hline
	\multirow{3}{*}{SPP} 
	& 8  & 71.4  & 77.7  & \textbf{85.1} & 61.1  & 54.7  & 64.2  & 85.5 \\
	& 16 & 70.0  & 71.8  & 85.0  & 62.5  & 55.3  & 62.3  & 82.8 \\
	& 32 & 77.3  & 79.7  & 83.0  & 70.9  & \textbf{67.3} & 75.3  & 87.4 \\
	\hline
\end{tabular}

\end{table*}

\section{Results}
In the following sections, we will present the results of the methods, and then provide an analysis of the implications arising from the involving factors in Discussion section.  

\subsection{Comparison of Proposed Methods}

To evaluate the effectiveness of the proposed methods introduced in Section~\ref{sec:decompose}, we conducted a series of few-shot learning experiments. Each task was evaluated under 8-, 16-, and 32-shot settings, and all experiments were repeated three times with different random seeds to ensure result reliability.

\added[id=R3,comment={C \#5,  C\#2 Comparison to other  																																														methods}]{
In addition to our proposed methods—\textbf{SPP} and \textbf{SCP}, we include two baseline comparisons}:

\begin{itemize}
	\item \textbf{PT (Prompt Tuning)}: A single-task baseline where each task uses its own private prompt trained independently.
	\item \textbf{MST (Multi-task Shared Prompt Tuning)}: A multi-task baseline where all tasks share a single source prompt without any private component.
\end{itemize}

Tables~\ref{table:res} and~\ref{table:res_lt} report the average performance of all methods across the GLUE and SET2 benchmarks.

\subsubsection{Few-shot Efficiency} 
\added[id=R3,comment={C \#2 Advatage over prompt tuning}]
{Across all few-shot settings (8-, 16-, and 32-shot), the proposed methods—SPP and SCP—consistently outperform the baselines. For instance, SPP achieves an average accuracy of 81.0\% at the 32-shot level, compared to 70.15\% for PT. This trend is observed across both GLUE and SET2 benchmarks and across all evaluated sample sizes.}

The proposed methods also surpass MST, demonstrating superior performance compared to using a shared source prompt alone. A more detailed analysis of this observation is provided in the Discussion section (Section~\ref{sec:discuss}).

Performance gains over prompt tuning are particularly pronounced in tasks such as QNLI, RTE, and MNLI within the GLUE benchmark, and PAWS, MNLI, and SciTail in the SET2 benchmark.

\begin{table*}[th!]
	\centering
	
	\caption{Comparison of proposed methods on 5000 Data Samples for GLUE Tasks}
	\label{table:res_gt_5k}
	\begin{tabular}{llrlllllllll}
\hline
 label   & Avg.           &   nsp    & cola          & mnli          & mrpc          & qnli          & qqp           & rte           & sst2          & stsb          \\
\hline
 SPP    & \textbf{86.01} &                   10             & \textbf{81.8} & 82.1          & 86.0          & \textbf{92.8} & 89.6          & 73.3          & \textbf{93.2} & \textbf{89.3}         \\
  SCP    & 85.89          &                    3             & 80.5          & \textbf{83.1} & \textbf{87.0} & 92.7          & 89.6          & 72.2          & 93.0          & 89.0                 \\
\hline
\end{tabular} \end{table*}
\begin{table*}[th!]
	\centering
	
	\caption{Comparison of proposed methods on 5000 Data Samples for SET2 Tasks}
	\label{table:res_lt_5k}
	\begin{tabular}{llrccccccc}
\hline
 label   & Avg.           &  nsp    & imdb          & mrpc          & multinli      & paws          & scitail       & yelp   \\
\hline
 SPP    & \textbf{88.53} &                   10             & \textbf{86.4} & 86.8          & \textbf{83.7} & \textbf{88.8} & 92.7          & 92.8            \\
  SCP    & 88.23          &                    3          & 86.0          & \textbf{87.2} & 82.9          & 88.2          & 91.9          & \textbf{93.2}   \\
\hline
\end{tabular} \end{table*}

\subsubsection{More Training data}
We have expanded our experiments to include a larger training dataset. The results of our proposed methods, utilizing 5000 samples for each task, are displayed in Table \ref{table:res_gt_5k} for GLUE tasks and Table \ref{table:res_lt_5k} for tasks in SET2.

While our designed methods demonstrate significant improvements in performance in few-shot settings, they also achieve state-of-the-art results with a larger training dataset, outperforming the overall performance compared to the related works, which are detailed in the Related Works section \ref{sec:relworks}.

\section{Discussion} \label{sec:discuss}
In this section, we explore the pivotal roles played by source and private prompts in our proposed methods, emphasizing the control and balance required to achieve desirable results. We delve into how these prompts contribute to task performance and knowledge transfer, examining strategies to optimize their impact for enhanced model performance.

\subsection{Role of Source and Private Prompt Combination} 

Figure \ref{fig:nsp} compares our proposed methods (SPP and SCP) with the baselines (PT and MST) across varying numbers of source prompts. Regardless of the number of source prompts, both SPP and SCP—which combine a private prompt with one or more source prompts—consistently outperform the baselines that rely solely on either a single source prompt or distinct private prompts for each task.

Interestingly, in the case of SET2, using a single source prompt outperforms the use of separate private prompts. This suggests a high degree of knowledge sharing among the tasks in this set, which primarily fall into three clusters: natural language inference (NLI), paraphrasing, and sentiment analysis with consistent class labels. In contrast, the opposite trend is observed in GLUE tasks, where using a single source prompt results in worse performance. This may be attributed to task conflicts arising from similar input structures but divergent objectives (e.g., CoLA vs. SST-2) or differing class labels (e.g., QQP vs. MRPC).

In both benchmarks, the use of private prompts in SPP and SCP enables task-specific adaptation, helping to resolve such conflicts and leading to consistent improvements over the baselines.

As shown in Figure~\ref{fig:nsp}, the SPP method outperforms SCP, due to its shorter prompt length and adaptive structure. By using a weighted combination rather than concatenation, SPP allows the model to balance and compensate between source and private prompts. In contrast, SCP requires both components to be independently strong, as their interaction is fixed and indirect. This makes SPP more efficient and robust in integrating complementary task-specific information.

\subsection{Number of Source Prompts} 

\begin{figure}
	\centering
	\includegraphics[width=0.48\linewidth]{ 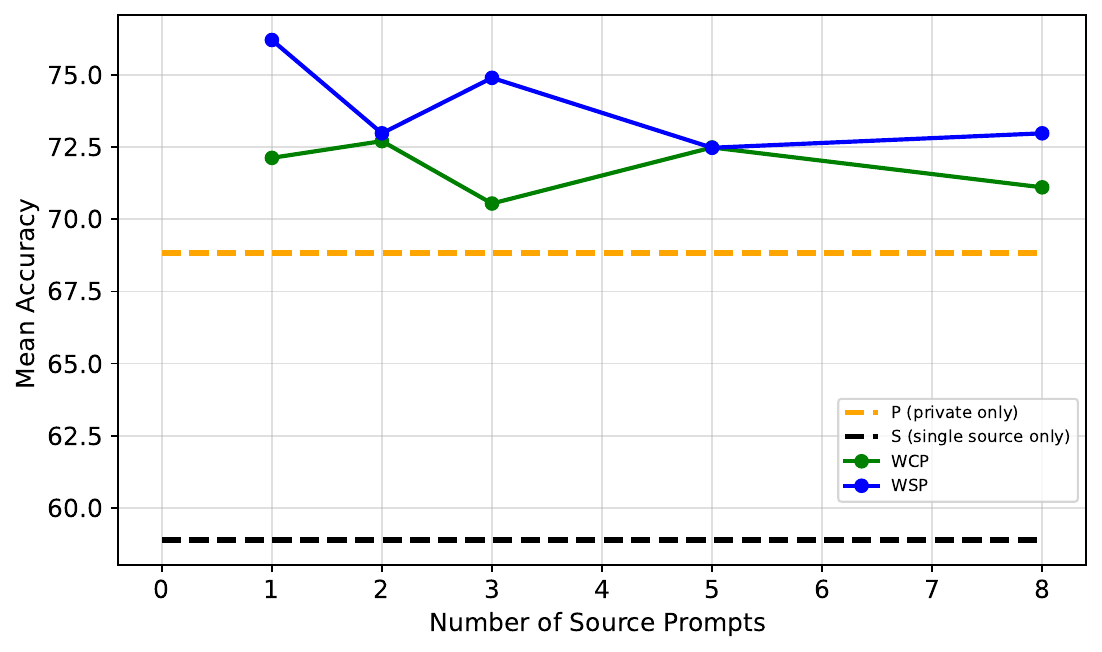}
    \includegraphics[width=0.48\linewidth]{ 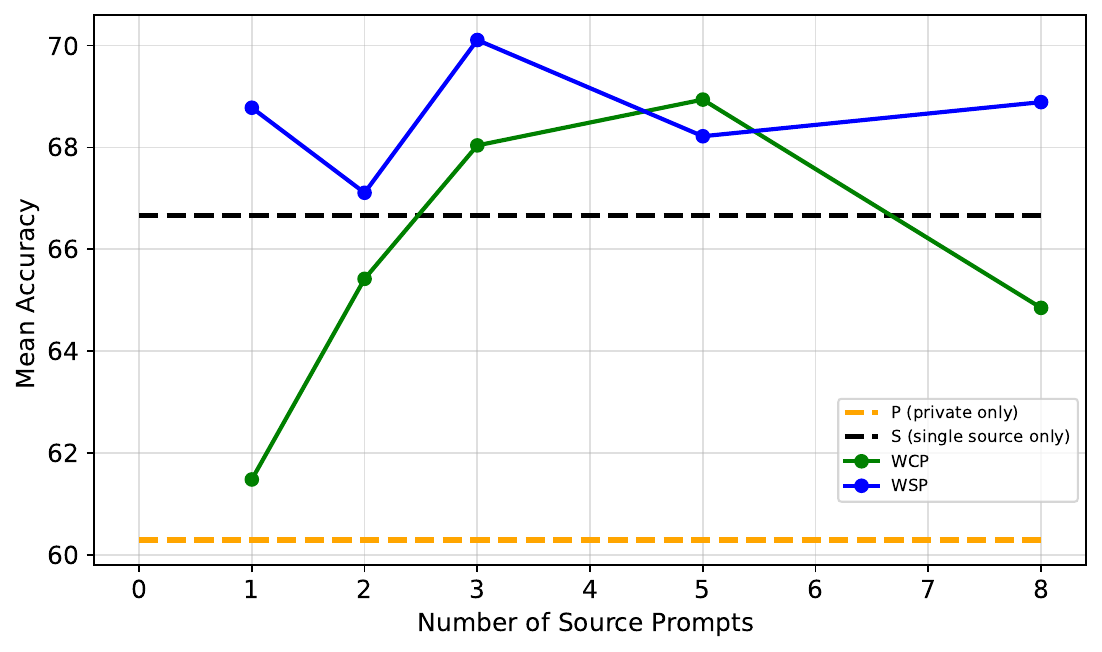}
	\caption{Average performance of the SPP method on GLUE tasks (left) and SET2 tasks (right), trained with 16 samples, across different numbers of source prompts.}
	\label{fig:nsp}
\end{figure}

Figure \ref{fig:nsp} also illustrates the impact of the number of source prompts on the performance of the proposed methods. While performance generally improves with fewer source prompts, SPP remains effective even when more source prompts are used.

\added[id=R3,comment={C \#4 Optimization challenges}]{
The optimal number of source prompts varies with task characteristics. As shown in Figure~\ref{fig:nsp}, GLUE tasks generally perform better with fewer—often a single—source prompt, whereas SET2 tasks benefit from using three or more.} This difference reflects the underlying task structure: SET2 tasks can be grouped into three distinct clusters with limited overlap, necessitating more diverse source knowledge. In contrast, despite differences in label space or objectives, GLUE tasks exhibit substantial shared structure. 
\added[id=R3,comment={C \#1 Balancing knowledge transfer}]{While private prompts help resolve superficial conflicts, a dense single source prompt can capture generalizable knowledge that benefits multiple tasks, thereby enhancing information sharing.}

\begin{figure}
	\centering
	\includegraphics[width=0.8\linewidth]{ 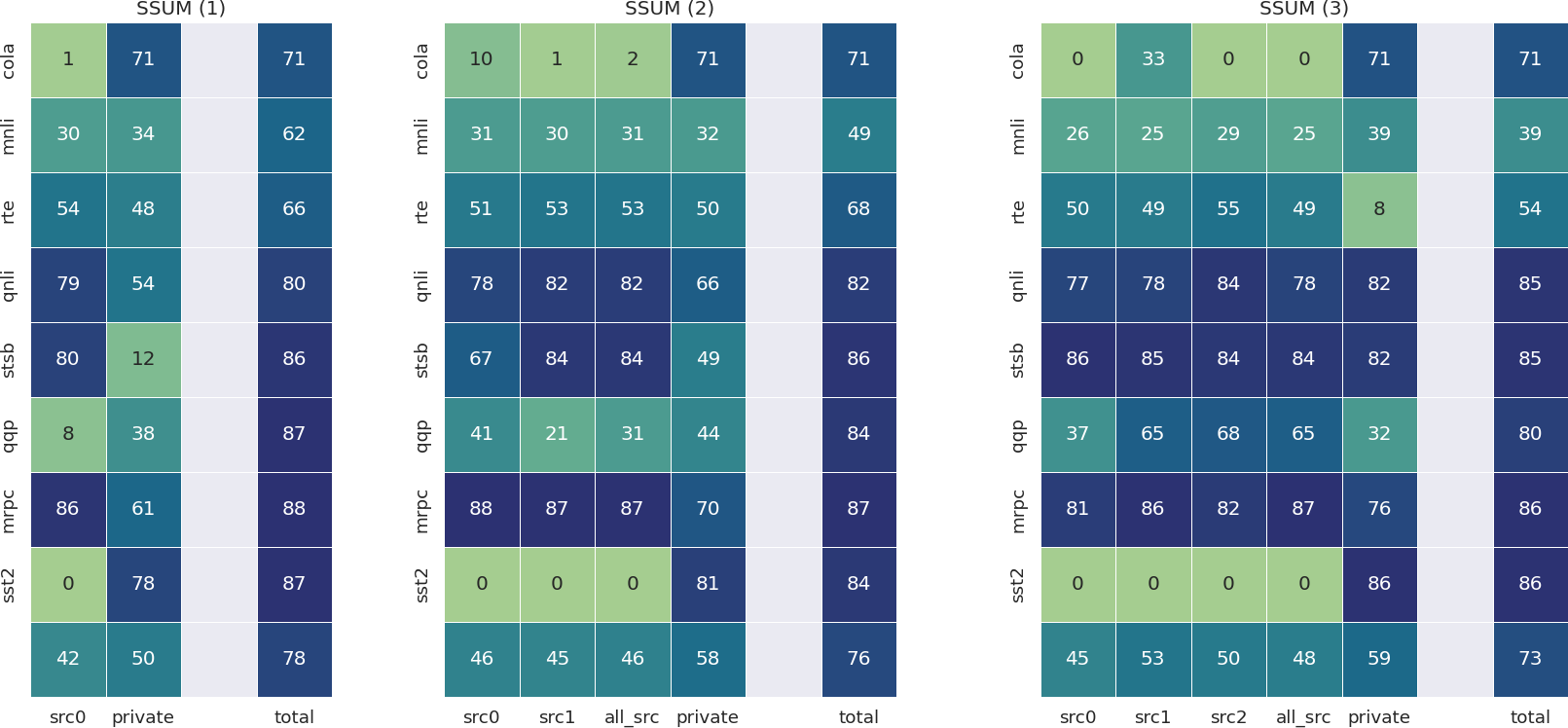}
	\caption{Performance comparison of the SPP method on GLUE tasks  trained with 16 samples using one, two, and three source prompts. Each column represents the performance when retaining a specific prompt and ignoring the others. The "all\_src" column ignores the private prompt, while "total" includes all prompts. The last row indicates the average score of each column.}
	\label{fig:SPP}
\end{figure}

\subsection{Contribution of Source Prompts} 
To analyze the impact of individual source prompts, we conducted ablation tests by isolating each source prompt and masking the rest, then evaluated performance on a 100-sample test set. The results for each proposed method are discussed below.

Figure \ref{fig:SPP} shows scores for the SPP method using one, two, and three source prompts across GLUE tasks. Each column reflects performance when only the corresponding source prompt is active; other source prompts and the private prompt are masked. The column \texttt{all\_src} shows results using all source prompts without the private prompt, and \texttt{private} shows results using only the private prompt. The final column \texttt{total} reports scores when both source and private prompts are used together.

Tasks like SST-2 and CoLA depend almost entirely on private prompts. These tasks differ most from others in both input format and output space, and—per Table \ref{table:res}—benefit least from multi-task prompt tuning.

Conversely, STS-B, QNLI, and MRPC rely heavily on source prompts, while tasks like QQP, RTE, and MNLI benefit from both prompt types. In these cases, combining source and private prompts improves performance beyond what either achieves alone, emphasizing their complementary roles. This is consistent with Table \ref{table:res}, where these tasks show the greatest gains from multi-task prompt tuning.

\begin{table*}[h!]
	\centering
	\caption{Predictions for the MNLI task using the source prompt, the private prompt, and their combination in the SPP method.}
	\label{table:mnli}
	\begin{tabular}{llllll}
\hline
 method                         & preds          & contradiction   & entailment   & neutral   & total acc.                          \\
\hline
\hline
 \multirow{3}{*}{source}   & contradiction  & 1               & 0            & 0       & \multirow{3}{*}{0.30}          \\
                                & \textbf{entailment}     & 27              & 29           & 36      &                                \\
                                & not-entailment & 4               & 0            & 2       &                                \\
 \rowcolor{gray!20}             & precision      & 3\%           & \textbf{100}\%      & nan       &                                \\
\hline
 \multirow{3}{*}{private} & \textbf{contradiction}  & 32              & 24         & 36        & \multirow{3}{*}{0.34}          \\
                                & neutral        & 0               & 5          & 2         &                                \\
 \rowcolor{gray!20}             & precision      & \textbf{100}\%         & nan          & 5\%     &                                \\
\hline
 \multirow{3}{*}{combined}  & contradiction  & 19              & 1            & 4         & \multirow{3}{*}{\textbf{0.62}} \\
                                & entailment     & 2               & 10           & 2         &                                \\
                                & neutral        & 11              & 18           & 32        &                                \\
 \rowcolor{gray!20}             & precision      & 59\%          & 34\%       & 84\%    &                                \\
\hline
\end{tabular} \end{table*}

\begin{table*}[h!]
	\centering
	\caption{Predictions for the RTE task using the source prompt, the private prompt, and their combination in the SPP method.}
	\label{table:rte}
	\begin{tabular}{lllll}
\hline
 method                         & preds          & entailment   & not-entailment   & total acc.                          \\
\hline
\hline
 \multirow{3}{*}{source}   & entailment     & 6            & 2                & \multirow{3}{*}{0.54}          \\
                                & \textbf{not-entailment} & 43           & 48               &                                \\
 \rowcolor{gray!20}             & precision      & 12\%       & \textbf{96}\%           &                                \\
\hline
 \multirow{1}{*}{private} & \textbf{entailment}     & 49           & 50             & \multirow{1}{*}{0.48}          \\
 \rowcolor{gray!20}             & precision      & \textbf{100}\%      & nan              &                                \\
\hline
 \multirow{3}{*}{combined}  & entailment     & 43           & 27               & \multirow{3}{*}{\textbf{0.66}} \\
                                & not-entailment & 6            & 23               &                                \\
 \rowcolor{gray!20}             & precision      & 88\%       & 46\%           &                                \\
\hline
\end{tabular} \end{table*}

\paragraph{Complementarity of Prompts on MNLI and RTE}
\added[id=R3,comment={C \#3 Effect of source prompt choice}]{We further examined how predictions vary depending on which prompt is used.} Table \ref{table:mnli} shows that, for MNLI, the source prompt strongly favors the \texttt{entailment} class—shared with QNLI and RTE, which dominate the source space (see Figure \ref{fig:SPP}-1). In contrast, the private prompt primarily predicts \texttt{contradiction}. Their combination enables the model to predict all three MNLI classes, achieving a significantly higher accuracy. This illustrates how the source and private prompts capture complementary semantics.

For RTE (Table \ref{table:rte}), the pattern is reversed: the source prompt emphasizes \texttt{not-entailment}, while the private prompt predicts \texttt{entailment}. This contrast enhances RTE-MNLI differentiation in the shared source space.

Table \ref{table:qqp} provides the predictions for QQP task using only the source prompt,  only the private prompt, and their combination in the SPP method. 
\begin{table*}[bh!]
	\centering
	\caption{Predictions for the QQP task using the source prompt, the private prompt, and their combination in the SPP method.}
	\label{table:qqp}
	\begin{tabular}{lllll}
		\hline
		method                         & preds          & duplicate   & not-duplicate   & total acc.                          \\
		\hline
		\hline
		\multirow{3}{*}{source}   & duplicate      & 1           & 0               & \multirow{3}{*}{0.8}           \\
		& entailment     & 7           & 0               &                                \\
		& not-duplicate  & 0           & 7               &                                \\
		& \textbf{not-entailment} & 30          & 54             &                                \\
		\rowcolor{gray!20}             & precision      & 3.0\%       & 11.0\%          &                                \\
		\hline
		\multirow{3}{*}{target} & \textbf{duplicate}      & 38          & 33              & \multirow{3}{*}{0.38}          \\
		& not            & 0           & 1               &                                \\
		& not-           & 0           & 4               &                                \\
		& not-duplicate  & 0           & 6               &                                \\
		& not-uplicate   & 0           & 17              &                                \\
		\rowcolor{gray!20}             & precision      & 100.0\%     & 10.0\%          &                                \\
		\hline
		\multirow{3}{*}{combined}  & duplicate      & 38          & 13              & \multirow{3}{*}{\textbf{0.87}} \\
		& not-duplicate  & 0           & 48              &                                \\
		\rowcolor{gray!20}             & precision      & 100.0\%     & 79.0\%          &                                \\
		\hline
\end{tabular} \end{table*}

While QQP is structurally similar to QNLI, it exhibits a form of label-space conflict, as the source prompt—optimized across tasks—tends to predict the \texttt{not-entailment} class for QQP (see Table \ref{table:qqp}). This label, while originating from entailment tasks, aligns more closely with QQP's \texttt{not-duplicate} class in practice. To counterbalance this misalignment, the private prompt for QQP predominantly predicts the \texttt{duplicate} class, thereby compensating for the shared prompt's bias. As a result, the combined prompt achieves better performance than either component alone, suggesting an effective integration mechanism that adapts shared knowledge (from QNLI) to the specific needs of QQP. 
\added[id=R3,comment={C \#3 Effect of source prompt choice}] {This highlights how QQP can still benefit from QNLI in the shared space, despite their distinct labeling schemes.}

\begin{figure}
	\centering
	\includegraphics[width=0.4\linewidth]{ 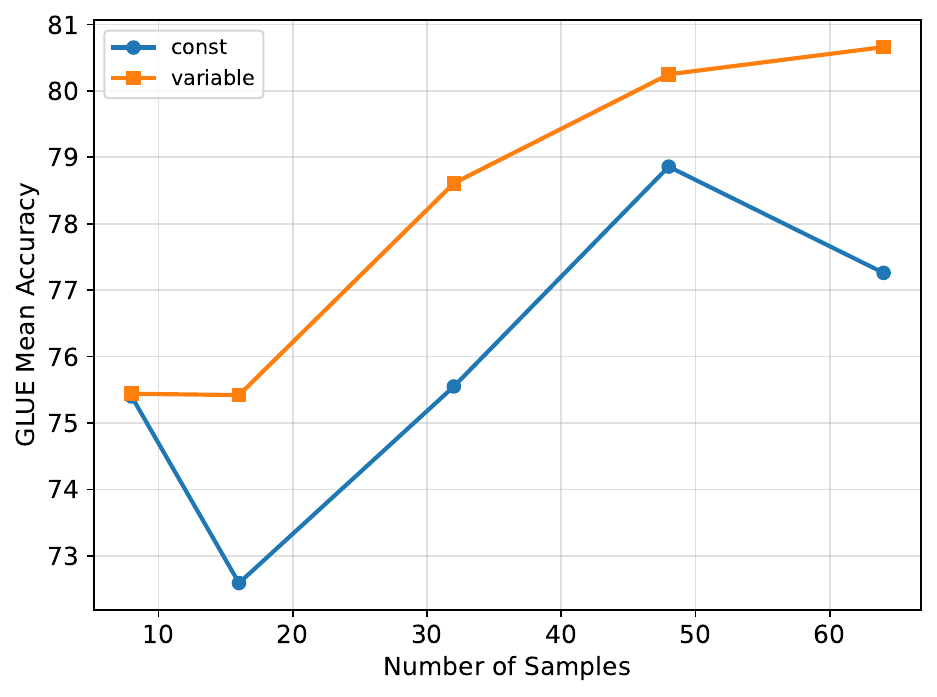}
	\includegraphics[width=0.4\linewidth]{ 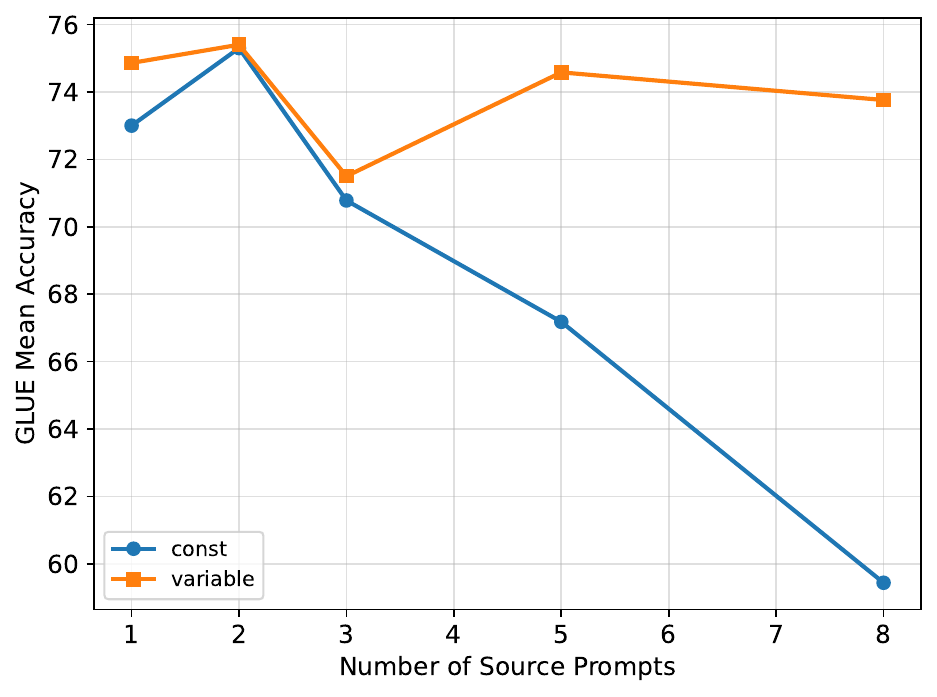}
	\includegraphics[width=0.4\linewidth]{ 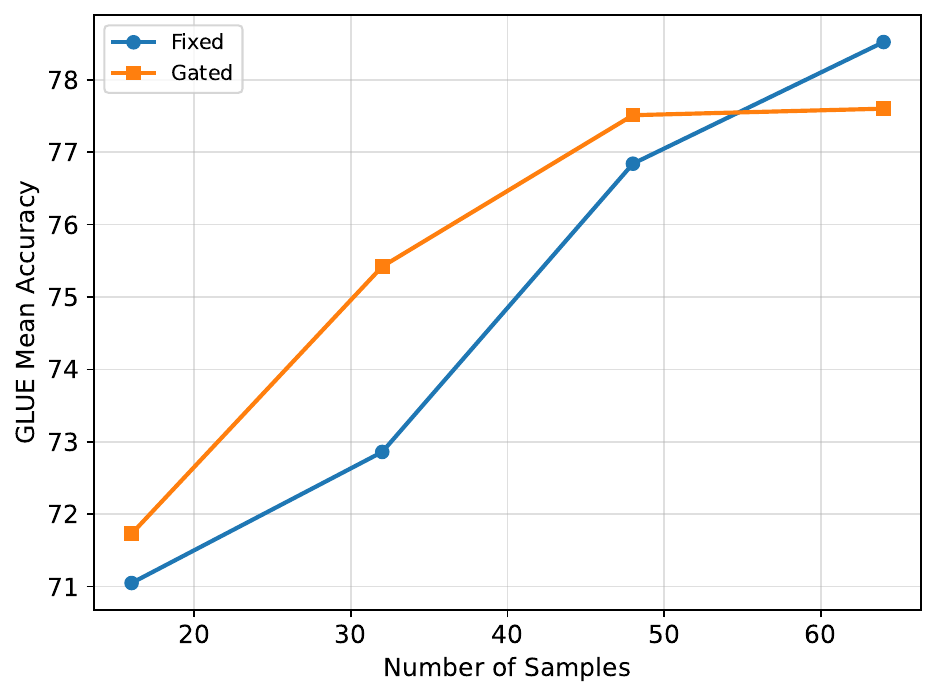}		
	\includegraphics[width=0.4\linewidth]{ 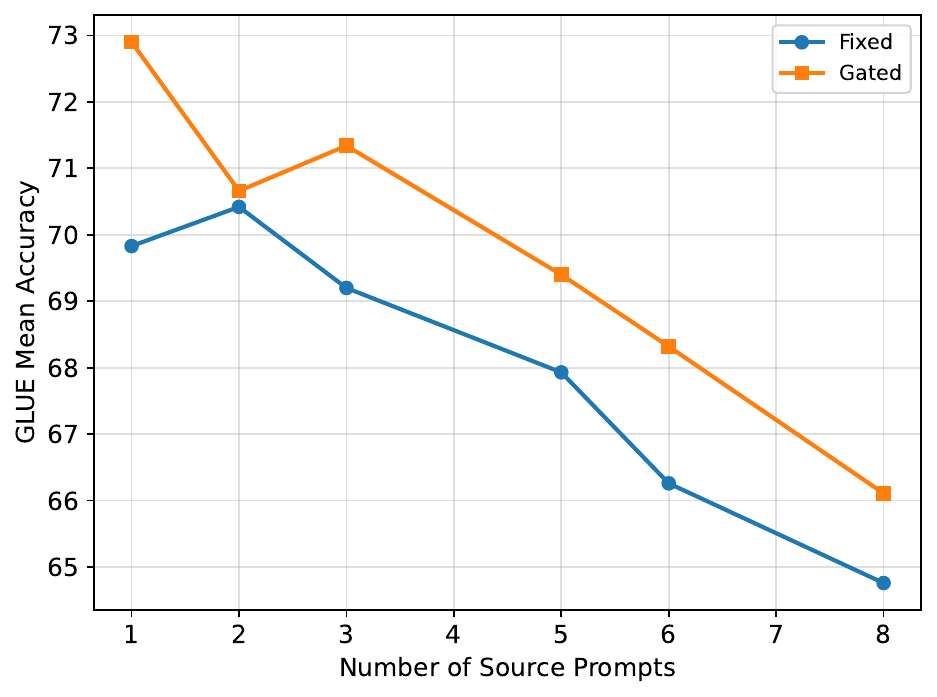}		
	\caption{The average performance of the proposed methods using constant and variable weights on GLUE tasks is shown across different numbers of training samples (top left) and source prompts (top right), using 16 samples. The effect of the gating coefficients ($\alpha$ and $\beta$), which control the contribution of private and shared source prompts, is shown by comparing performance with and without them across various numbers of samples (bottom left) and source prompts (bottom right).}
	\label{fig:const}
\end{figure}

\subsection{The Impact of Attention Module}
An important question is whether the attention module is necessary—what happens if we instead assign uniform (constant) weights to all source prompts? Figure~\ref{fig:const} compares the performance of using fixed constant weights versus learned variable weights, under varying numbers of source prompts and data samples. To analyze the effect of sample size independently, we fixed the number of source prompts at two while varying the number of data samples.

We observe that constant weights perform comparably when training data is limited. \added[id=R3,comment={C \#1 Balancing knowledge transfer}] { However, as more data becomes available, learned weights yield superior results. }This suggests that in few-shot scenarios, the available data may be insufficient to train the attention module effectively, but in richer data settings, learning source prompt weights becomes increasingly beneficial.

\added[id=R3,comment={C \#1 Balancing knowledge transfer}]{
We further investigated how the weighting mechanism behaves as the number of source prompts increases.} Our hypothesis was that learned weights could help differentiate the roles of source prompts and mitigate potential task interference. As illustrated in Figure~\ref{fig:const}, learning weights has a pronounced impact on performance, particularly as the number of source prompts increases.

\begin{figure}
	\centering
	\includegraphics[width=0.7\linewidth]{ 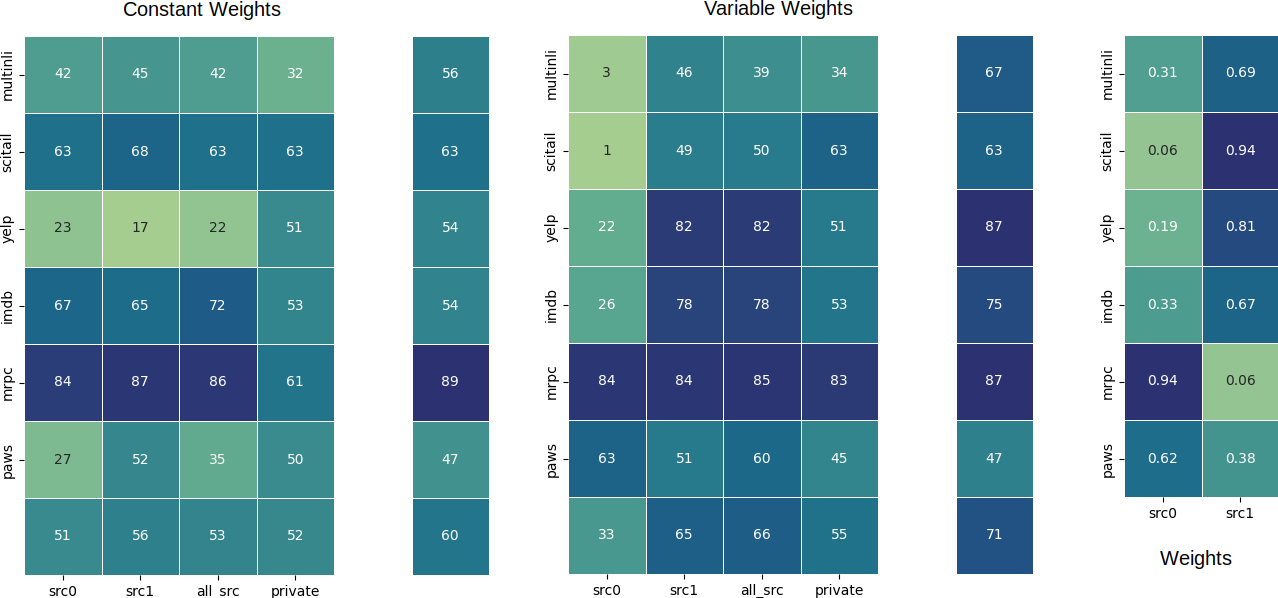}
	\caption{Analysis of individual prompt scores for tasks in SET2 with constant weights (left) and variable weights (right), alongside corresponding learned weights.}
	\label{fig:const-vs-var-lt}
\end{figure}

To highlight this behavior, we analyzed the isolated effects of individual prompts. Figure~\ref{fig:const-vs-var-lt} presents results for SET2 when using two source prompts under constant and variable weighting schemes. With constant weighting, all sources contribute equally, which can obscure task-specific distinctions. In contrast, variable weights enable more precise attribution of influence, effectively disentangling related task groups—such as MRPC vs. PAWS (paraphrasing) and Yelp-Polarity vs. IMDB (sentiment)—and thereby improving performance on each.

\added[id=R3,comment={C \#1 Balancing knowledge transfer}]{
Furthermore, we examined the impact of the gating coefficients $\alpha$ and $\beta$ in the SPP method, which control the contributions of source and private prompts to the target prompt.} As shown in the bottom panels of Figure~\ref{fig:const}, we compare a Fixed configuration ($\alpha=\beta=1$) with the Gated setup, where $\alpha$ and $\beta$ are learned. The results demonstrate that learning these coefficients consistently improves performance across varying numbers of training samples and source prompts—particularly in few-shot settings, where adaptive weighting enables better utilization of limited task-specific signals.

However, when training data is abundant, the fixed weighting performs slightly better. This may be because, in high-data regimes, the model can learn effective representations directly from the task data, reducing the need for dynamic gating. Fixed coefficients may also provide more stable optimization by eliminating the need to learn additional parameters.

\begin{figure}[th!]
	\centering
	\includegraphics[width=0.45\linewidth]{ 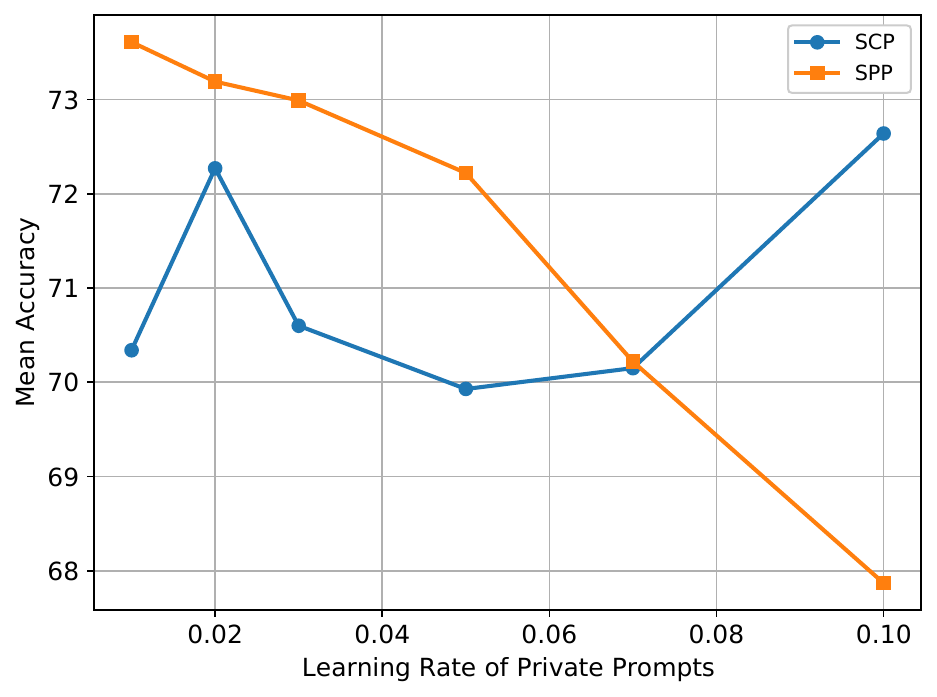}
	\caption{Performance comparison of the proposed methods across different learning rates for the private prompt on GLUE tasks using 16 samples.}
	\label{fig:lr}
\end{figure}

\subsection{Balancing the Impact of Source and Private Prompts}

To optimize performance, we explore the effect of balancing source and private prompts by assigning them different learning rates. Figure~\ref{fig:lr} illustrates the performance of our methods across varying learning rates for the private prompt, while keeping the learning rate of the source prompt fixed at 0.05 (measured at 30 epochs using 16 samples per task).

\added[id=R3,comment={C \#1 Balancing knowledge transfer}]{
Adjusting the relative learning rates allows us to control the model’s reliance on the source versus private prompts. In general, better performance is observed when the source prompt uses a higher learning rate than the private prompt. This suggests that leveraging generalizable knowledge from source prompts is beneficial, especially in early training.}

As seen previously, tasks like MNLI, RTE, and QQP benefit significantly from the complementary roles of source and private prompts. This trend is particularly evident in the SPP method. When the private prompt’s learning rate is low, the model leans more on the source prompts, achieving stronger performance. As the private prompt’s learning rate increases, the model begins to shift focus toward the private prompt. However, in intermediate ranges, neither prompt dominates, leading to degraded performance—likely because the model fails to fully exploit either source. 

\added[id=R3,comment={C \#4 Optimization challenges}]{
When the private prompt’s learning rate becomes large, performance partially recovers, as the model increasingly relies solely on the private prompt. This effect is particularly pronounced in SCP, where the private prompt is not blended with source prompts and can therefore dominate without interference. In contrast, for SPP, increasing the private prompt’s learning rate degrades performance—likely due to insufficient task-specific data, which limits the private prompt’s capacity to generalize effectively. This highlights how the model leverages shared examples across tasks via the source prompt to boost performance in low-resource settings.}

\added[id=R3,comment={C \#1 Balancing knowledge transfer}]{
These observations underscore the importance of learning rate scheduling as a mechanism to balance general knowledge from source prompts with task-specific adaptation from private prompts.}

\begin{table*}[th!]
	\centering
	\caption{Comparison of proposed Methods on SET2 tasks by including MNLI as additional task}
	\begin{tabular}{lllllllll}
		\hline
		label & All & include & imdb & mrpc & multinli & paws & scitail & yelp \\
		\hline
		\multirow{2}{*}{SPP} & 72.7 & mnli & 77.0 & 88.0 & 65.0 & 49.0 & 70.0 & 87.0 \\
		& 70.5 & --- & 75.0 & 89.0 & 61.0 & 51.0 & 64.0 & 83.0 \\
		\midrule
		\multirow{2}{*}{SCP} & 74.8 & mnli & 83.0 & 88.0 & 72.0 & 53.0 & 68.0 & 85.0 \\
		& 71.3 & --- & 81.0 & 88.0 & 58.0 & 47.0 & 67.0 & 87.0 \\
		\midrule
		PT & 65.7 & --- & 73.0 & 52.0 & 61.0 & 59.0 & 65.0 & 84.0 \\
		\hline
	\end{tabular}
	\label{table:inc}
\end{table*}

\subsection{Modularity}

In our approach, the selection of tasks for multi-task prompt tuning plays a crucial role in determining performance. Additionally, our method supports initializing source prompts with pretrained prompts obtained from earlier training on individual tasks, reinforcing its modular design.
\added[id=R3,comment={C \#2 Advantage over prompt tuning}]{
This modularity allows for flexible inclusion or exclusion of source tasks to optimize performance on the target task. } \added[id=R3,comment={C \#3 Effect of source prompts choice}] {For instance, within the GLUE benchmark, excluding SST-2 as a source task can be justified due to its limited contribution to other tasks. In contrast, tasks such as MNLI or STS-B have shown strong utility and provide substantial benefit when included as source tasks.
As an illustrative example, we added MNLI to the task set of SET2 and compared performance with and without it. The results, shown in Table~\ref{table:inc}, demonstrate that the inclusion of MNLI improves performance across all methods—particularly for tasks like MultiNLI and SciTail, which are semantically close to MNLI.
Furthermore, due to the presence of private prompts in our method, tasks that do not benefit from one another can rely primarily on their private prompt. This design ensures that multi-tasking does not degrade performance, and in most cases, outperforms vanilla prompt tuning, as evidenced in Table~\ref{table:inc}.}

\section{Related Work} \label{sec:relworks}

This work falls within the broader area of parameter-efficient transfer learning for pretrained language models (PLMs). These methods aim to adapt PLMs to downstream tasks by fine-tuning only a small subset of parameters. The fine-tuned parameters may take the form of lightweight neural adapters inserted between transformer layers \cite{concept-tasklayer2}, soft prompts appended to input embeddings \cite{lester2021power}, modifications to hidden states \cite{li2021prompting}, learnable bias terms within model weights \cite{zaken2022bias}, or low-rank updates to attention and feedforward weights \cite{hu2021language}.

\added[id=R3,comment={C \#5 Comparison to other methods}]{
Our work is most closely related to prompt tuning approaches—specifically, multi-task prompt tuning—with an emphasis on modularity, single-stage training, and explicit compositionality of source and task-specific knowledge through soft prompts.}

\vspace{1ex}
\paragraph{Prompt Transfer and Initialization.}
Vu et al. \cite{spot} proposed SPoT, which investigates transferring prompt knowledge across tasks. In one variant, SPoT learns a shared prompt over multiple source tasks via multi-task training and then uses it to initialize the target task’s prompt. In another variant, SPoT learns separate prompts for each source task and selects the most similar one to initialize the target. Although SPoT improves transfer, it remains a two-stage process: pretrain prompts on source tasks, then transfer to target tasks.

Wang et al. \cite{rel-multi-pt} presented MPT (pretrained Prompt Tuning), which learns a shared prompt across tasks using decomposition and distillation. It also incorporates low-rank updates to adapt the shared prompt to specific tasks. Like SPoT, MPT uses a two-stage framework and focuses on shared representation learning.

\added[id=R2, comment={C \#2 Related works on vision}]{
In parallel, Shen et al. \cite{mvpt} introduced MVLPT (Multitask Vision-Language Prompt Tuning) in the vision domain. MVLPT explores multi-task prompt sharing in vision-language models and demonstrates that learning a transferable prompt from multiple source tasks can improve target task performance. It also investigates joint training with shared prompts across tasks and provides empirical insights into task compatibility and cross-task benefits. 
}
\added[id=R3,comment={C \#5 Comparison to other methods}]{
In contrast, our method eliminates the need for pretraining prompts separately. Instead, source prompts and private (task-specific) prompts are trained jointly in a single stage, with task prompts composed dynamically through various mechanisms, including weighted averaging, concatenation, and gated fusion.}

\vspace{1ex}
\paragraph{Modular Prompting.}
Sun et al. \cite{rel-mpm} proposed Multi-task Pretrained Modular Prompt ($MP^2$), which combines modular prompts with a trainable routing network. Their approach uses two stages: (1) training a pool of modular prompts with multi-task supervision and (2) selecting a subset for each target task. While $MP^2$ introduces modularity, the routing mechanism requires learning separate control parameters, and the modular prompts are frozen after pretraining.

Our method shares the goal of modularity but diverges by:
(1) Avoiding explicit routing, instead using learned attention or soft combinations of source prompts;
(2) Jointly training both source and private prompts in a single stage;
(3) Allowing task prompts to selectively leverage shared and private information during optimization. This flexible modularity ensures robustness, even when source tasks are weakly related.

\vspace{1ex}
\paragraph{Instance-wise Prompt Composition.}
ATTEMPT \cite{rel-attempt} by Asai et al. proposes composing task prompts from a set of pre-obtained source prompts using an instance-level attention mechanism. Their method learns a prompt for each target example by attending to the fixed source prompts and combining them with a randomly initialized prompt. While this dynamic attention offers flexibility, it introduces instance-wise overhead and relies on frozen source prompts, which limits adaptability—especially in few-shot scenarios.

By contrast, our approach supports global task-wise prompt composition, where attention weights or composition strategies are learned jointly with the source and private prompts. We demonstrate that fine-tuning source prompts—rather than freezing them—yields improved performance and stronger transfer, particularly under limited supervision.

\vspace{1ex}
\paragraph{Generative vs. Discriminative Settings.}
\added[id=R2,comment={C \#5. Alternative Fine Tuning Approaches }]{
A further point of differentiation is our focus on generative models for prompt tuning. Unlike many prior works that rely on classification-based fine-tuning (e.g., linear probing), our framework accommodates both classification and regression tasks in a generative formulation. This is particularly significant, as prompt tuning in generative models requires the model to recover and align with pretrained linguistic knowledge to generate suitable target outputs \cite{rethink}}.

While this approach offers broader task coverage, it also introduces unique challenges—including label conflicts, and label formatting issues. These considerations are often overlooked in classification-centric methods but are critical to performance and stability in generative setups. Our results indicate that, despite these challenges, the generative setting remains competitive and offers promising opportunities for unified modeling.

\begin{table*}[t!]
	\centering
	\caption{Comparison with the related works on GLUE tasks}
	\label{table:related}
	\begin{tabular}{lccccccccl}
		\hline
		\textbf{method/ task} & \textit{mnli} & \textit{qqp} & \textit{qnli} & \textit{sst2} & \textit{stsb} & \textit{mrpc} & \textit{rte} & \textit{cola} & \textbf{avg.} \\
		\hline
		
		\textbf{ComPT-SPP} & 82.1 & 89.6 & 92.8 & 93.2 & 89.3 & 86.0 & 73.3 &\textbf{81.8} & \textbf{86.0} \\
		
		\textbf{FineTuning} & \textbf{86.8 }& \textbf{91.6} & 93.0 & 94.6 & 89.7 & 90.2 & 71.9 & 61.8 & 84.9\\
		\textbf{PT} & 81.3 & 89.7 & 92.8 & 90.9 & 89.5 & 68.1 & 54.7 & 10.6 & 72.2 \\
		
		\textbf{Spot} & 85.4 & 90.1 & 93.0 & 93.4 & 90.0 & 79.7 & 69.8 & 57.1 & 82.3 \\
		
		\textbf{ATTEMPT} & 84.3 & 90.3 & 93.0 & 93.2 & 89.7 & 85.7 & 73.4 & 57.4 & 83.4 \\
		
		\textbf{ATTEMPT-m} & 83.8 & 90.0 & \textbf{93.1} & \textbf{93.7} & \textbf{90.8} & 86.1 & 79.9 & 64.3 & 85.2 \\
		
		\textbf{MPT} & 84.3 & 90.0 & 93.0 & 93.3 & 90.4 & 89.2 & \textbf{82.7 }& 63.5 & 85.8 \\
		\hline
	\end{tabular}
\end{table*}

\subsection{Comparison with Related Works}
In Table \ref{table:related}, we conducted a comparison between our method and the reviewed related works, which include vanilla prompt tuning (PT) \cite{lester2021power}, SPOT, ATTEMPT, and MPT, across 8 GLUE tasks. 'ATTEMPT-m' refers to the multi-tasking version of ATTEMPT. The reported numerical values are directly cited from the respective published papers. It's worth noting that $\text{MPT}^2$ evaluated their approach on Chinese tasks, which led us to exclude them from this particular comparison.

For our comparison, we utilized the test sets of the tasks. However, due to resource constraints, we conducted testing on a subset of 5000 samples for our best-performing configurations. Despite this limitation, our overall performance outperforms these other methods, and our scores are close to the top scores for various tasks. Nevertheless, we consider the main advantage of our approach to be in few-shot settings, where it boosts the performance of tasks through knowledge sharing among related tasks. 

\section{Conclusion}

In this work, we explored modular multi-task prompt tuning using shared source prompts and task-specific private prompts. Our findings show that combining these components—along with attention-based weighting and learning rate balancing—enables effective knowledge transfer across tasks, especially in few-shot settings. Our ablation studies confirm that each component—source prompts, private prompts, attention weighting, and learning rate tuning—contributes meaningfully to overall performance.

The proposed single-stage approach improves efficiency and flexibility by learning all prompts jointly, without requiring separate pretraining. Our method supports classification and regression tasks using generative models, highlighting both the benefits and challenges of this setting.

In general, our framework offers a unified and flexible approach to multi-task prompt tuning, combining the benefits of modularity, parameter efficiency, and transferability—while simplifying the training pipeline and extending applicability to a broader range of tasks. Future work could build on these findings by exploring more complex generative tasks in NLP or applying similar compositional prompt designs to tasks in computer vision.

\section*{Declarations}

\begin{itemize}
	\item \textbf{Competing Interests:} The authors declare no competing interests.
	\item \textbf{Authors' Contributions:}
	\begin{itemize}
		\item Hesham Faili: Supervision, Conceptualization, Methodology, Resources, Validation, Writing - Review \& Editing.
		\item Ahmad Pouramini: Conceptualization, Investigation, Data Curation, Methodology, Writing - Original Draft Preparation.
	\end{itemize}
	\item \textbf{Ethical and Informed Consent for Data Used:} This study did not involve human or animal subjects. All data are publicly available and cited appropriately. Informed consent was not required.
	\item \textbf{Data Availability and Access:} Data are available on a public repository. Code will be  publicly available on GitHub: \url{https://github.com/puraminy/ComPT}
\end{itemize}

\bibliography{sn-bibliography}
\clearpage

\end{document}